\documentclass[letterpaper, 12 pt, conference]{ieeeconf}  

\IEEEoverridecommandlockouts                              
\usepackage{graphicx} 
\usepackage{bm}

\usepackage{epsfig} 
\usepackage{mathptmx} 
\usepackage{times} 
\usepackage{amsmath} 
\usepackage{amssymb}  
\usepackage{subcaption}

\usepackage[pagebackref=true,breaklinks=true,letterpaper=true,colorlinks,bookmarks=false]{hyperref}

\begin{document}

\title{DeepVO: A Deep Learning approach for Monocular Visual Odometry}

\author{Vikram Mohanty \hspace{0.2cm} Shubh Agrawal \hspace{0.2cm} Shaswat Datta \hspace{0.2cm} Arna Ghosh\\Vishnu D. Sharma \hspace{0.2cm} Debashish Chakravarty\\
Indian Institute of Technology Kharagpur\\
Kharagpur, West Bengal, India 721302\\
{\tt\small \{vikram.mohanty, shubh.agrawal111, shaswatdatta, arna.ghosh, vds,  dc\}@iitkgp.ac.in}
}

\maketitle

\begin{abstract}
Deep Learning based techniques have been adopted with precision to solve a lot of standard computer vision problems, some of which are image classification, object detection and  segmentation. Despite the widespread success of these approaches, they have not yet been exploited largely for solving the standard perception related problems encountered in autonomous navigation such as Visual Odometry (VO), Structure from Motion (SfM) and Simultaneous Localization and Mapping (SLAM). This paper analyzes the problem of Monocular Visual Odometry using a Deep Learning-based framework, instead of the regular 'feature detection and tracking' pipeline approaches. Several experiments were performed to understand the influence of a known/unknown environment, a conventional trackable feature and pre-trained activations tuned for object classification on the network's ability to accurately estimate the motion trajectory of the camera (or the vehicle). Based on these observations, we propose a Convolutional Neural Network architecture, best suited for estimating the object's pose under known environment conditions, and displays promising results when it comes to inferring the actual scale using just a single camera in real-time. 
   
\end{abstract}

\section{Introduction}

In recent years, Convolutional Neural Networks (CNNs) have been employed successfully for numerous applications in Computer Vision and Robotics such as object detection \cite{ren2015faster} , classification \cite{krizhevsky2012imagenet}, semantic segmentation \cite{long2015fully} and many others, often outperforming the conventional feature-based methods. However, a few exceptions exist to this trend; notably - Structure from Motion (SFM), Simultaneous Localization and Mapping (SLAM) and Visual Odometry (VO) are some of the traditional perception problems, for which deep learning techniques have not been exploited in a large manner. In this paper, we analyze the problem of Visual Odometry using a Deep Learning-based framework.

In robot navigation, odometry is defined as the process of fusing data from different motion sensors to estimate the change in the robot's position over time. This process of determining the trajectory plays an important part in robotics, forming the basis of path planning and controls. Traditionally, this problem has been tackled using data from rotary encoders, IMU and GPS \cite{moore2016generalized}. While this approach has been practically successful in solving the problem in hand, it is still prone to unfavorable conditions like wheel slipping in uneven terrains and lack of GPS signals. Recently, this problem has been solved just by using data from the camera (sequence of images). This process of incrementally estimating the robot's pose (position and orientation) by analyzing the motion changes in the associated camera images is known as visual odometry \cite{scaramuzza2011visual}.

A standard Visual Odometry approach generally follows the following steps (for both monocular and stereo vision cases) \cite{maimone2007two} :

\begin{enumerate}
\item Image acquisition at two time instances
\item Image correction such as rectification and lens distortion removal
\item Feature detection in the two images (such as corners using SURF \cite{bay2006surf}, ORB \cite{rublee2011orb} or FAST \cite{rosten2006machine})
\item Feature tracking between the two images to obtain the optical flow
\item Estimation of motion using the obtained optical flow and the camera parameters.
\end{enumerate}

On the deep learning front, there have been huge technological advancements regarding the applications of CNNs. It has been shown that these deep networks are adept in extracting various abstract features from images. 

Our work proposes a Deep Learning-based framework for analyzing the problem of visual odometry, motivated from the observation that instead of geometric feature descriptors, CNNs can be used to extract high-level features from images. Using these features, we estimate the transformation matrix between two consecutive scenes to recreate the vehicle's trajectory. Another significant contribution of this paper is using only monocular vision to estimate the vehicle's position in true scale, which cannot be done solely by pure geometry based methods. This is possible since the training network is able to learn the camera intrinsic parameters and scale. We hope that this framework will open up further research into the associated fields of Simultaneous Localization and Mapping (SLAM) and Structure from Motion (SFM) as well.

\section{Related Work}

\subsection{Visual Odometry}

The problem of visual odometry has been traditionally tackled by two methods - feature-based and direct ("appearance-based"). While the first approach relies on detecting and tracking a sparse set of salient image features such as lines and corners, the latter relies directly on the pixel intensity values to extract motion information.

Feature-based methods use a variety of feature detectors to detect salient feature points such as FAST (Features from Accelerated Segment Test) \cite{rosten2006machine}, SURF (Speeded Up Robust Features) \cite{bay2006surf}, BRIEF (Binary Robust Independent Elementary Features) \cite{calonder2010brief}, ORB (Oriented FAST and Rotated BRIEF) \cite{rublee2011orb} and Harris \cite{harris1988combined} corner detectors. These feature points are then tracked in the next sequential frame using a feature point tracker, the most common one being the KLT tracker \cite{tomasi1991detection}, \cite{shi1994good}. The result thus obtained is the optical flow, following which the ego-motion can then be estimated using the camera parameters as proposed by Nister \cite{nister2004efficient}. This general approach of detecting feature points and tracking them is followed by most papers (in both monocular vision and stereo vision based approaches) as is the case in \cite{matthies1989dynamic} and \cite{johnson2008robust}. More recent works in this area employ the PTAM approach \cite{klein2007parallel}, which is a robust feature tracking-based SLAM algorithm, with an added advantage of running in real-time by parallelizing the motion estimation and mapping tasks \cite{blosch2010vision}, \cite{weiss2013monocular}, \cite{kneip2011robust}. 

Direct or "appearance-based" methods for visual odometry rely directly on the pixel intensity values in an image, and minimize errors directly in sensor space, while subsequently avoiding feature matching and tracking. These methods however require a planarity assumption (e.g. homography). Early direct monocular SLAM methods like \cite{jin2003semi} and \cite{molton2004locally} make use of filtering algorithms for Structure from Motion, while in \cite{silveira2008efficient} and \cite{pretto2011omnidirectional} non-linear least squares estimation was used. Other approaches like DTAM \cite{newcombe2011dtam} compute a dense depth-map for each key-frame, which was used for aligning the whole image to find the camera pose. This is done by minimizing a global energy function. Since this approach is computationally intensive, heavy GPU parallelization is required. To mitigate this heavy computational requirement, the method described in \cite{engel2013semi} is proposed. Recently, fast direct monocular SLAM has also been achieved by the LSD-SLAM algorithm \cite{engel2014lsd}. 

Aside from these two approaches, the other notable method is a semi-direct approach to the problem, which combines the successful factors of feature-based methods (tracking many features, parallel tracking and mapping) with the accuracy and speed of direct methods. This was explored in the work by Scaramuzza et.al. \cite{forster2014svo} 

\subsection{Deep Learning Approaches}

With the advent of CNNs \cite{lecun1995convolutional}, numerous computer vision tasks have been solved very efficiently and with higher accuracy by these architectures as compared to traditional geometry-based approaches. Classification problems such as the ImageNet Large Scale Visual Recognition Competition (ILSVRC) \cite{ILSVRC15}, \cite{krizhevsky2012imagenet}, regression problems like depth regression \cite{eigen2014depth}, object detection \cite{ren2015faster} and segmentation problems \cite{long2015fully} have all been solved by these networks.

However, the domains of Structure from Motion, SLAM and Visual Odometry are still untouched by the advances in deep learning. Recently, optical flow between two images has been obtained by networks such as FlowNet \cite{fischer2015flownet} and EpicFlow \cite{revaud2015epicflow}. Homography between two images have also been estimated using deep networks in \cite{detone2016deep}. Nicolai, Skeele et al. applied deep learning techniques to learn odometry, but using laser data from a LIDAR\cite{nicolaideep}. The only visual odometry approach using deep learning that the authors are aware of the work of Konda and Memisevic \cite{konda2015learning}. Their approach however is limited to stereo visual odometry. Agrawal et al. \cite{agrawal2015learning} propose the use of ego-motion vector as a weak supervisory signal for feature learning. For inferring egomotion, their training approach treats the whole problem as a classification task. As opposed to this, we treat the visual odometry estimation as a regression problem. 

\section{Methodology}

The pipeline can be divided into two stages : Data Preprocessing and the CNN Framework, designed specifically for different experiments. 

\subsection{Data Preprocessing}

For our experiments, the KITTI Vision benchmark \cite{kitti} was used. The visual odometry dataset provided by KITTI consists of stereo-vision sequences collected while driving the vehicle in different environments. Since this work focuses on monocular vision, the video sequences collected from a single camera were considered. Of the 21 sequences available, 11 sequences with ground truth trajectories were used for training and testing sequences. These 11 sequences were further sorted into training and testing dataset, as per the need of our experiments. The original ground truth pose information is available in terms of a sequence of 3$X$4 transformation matrices which describe the motion of a vehicle between 0$^{th}$ time step to t$^{th}$ time step. These matrices were processed to generate the ground truth data in a new form describing the differential changes in translational motion ($\Delta$x, $\Delta$z, $\Delta\Theta$) of the vehicle, for all subsequent images in pairs I$_{t}$ and I$_{t+1}$  (where I$_{t}$ is image at t$^{th}$ time step and I$_{t+1}$ is image at (t+1)$^{th}$ time step) along two designated translational axes (x, z). Each of the original image sequences of size 1241$X$376 were warped and downsampled to 256$X$256, as the architecture we propose was inspired by AlexNet \cite{alexnet}, which restricts inputs to square sized images only. Later, a dataset of image pairs was generated consisting of images at t$^{th}$ time step and the corresponding image at (t+1)$^{th}$ time step. Thus, the final processed dataset could be represented as: \[
\left \{
\begin{tabular}{ccc} I$_{t}$ , I$_{t+1}$ , ($\Delta$x,$\Delta$z, $\Delta\Theta$)$_{t->(t+1)}$ \end{tabular} 
\right \} 
\]

This was the base input image and ground truth label format. However, for different experiments, this base data was converted into other realizable formats, or augmented with additional data, which are explained in the later subsections.

\subsection{Hardware and Software}

All the demonstrated experiments were performed on an  Intel Xeon @4 x 3.3 GHZ machine loaded with 32 GB DDR3 RAM and NVIDIA GTX 970. To evaluate our approach for learning visual odometry and GPU based implementations, we chose Caffe\cite{caffe}, developed by the Berkeley Vision and Learning Center. All the data pre-processing were programmed in Python, using associated libraries for compatibility with the python bindings of Caffe.

\subsection{Deep Learning Framework}

We designed a CNN architecture, partly based on the original AlexNet \cite{alexnet}, tuned to take as inputs simultaneously - the paired images in sequence (I$_{t}$, I$_{t+1}$), with an objective to regress the targeted labels ($\Delta$x, $\Delta$z, $\Delta\Theta$). All weights in the network's convolutional layers had a gaussian initialization, whereas the fully connected layers were initialized using the xavier algorithm \cite{glorot2010understanding}. The network was designed to compute a L2 (Euclidean) Loss. Based on the different experiments performed for the proposed analysis, the network architecture was further tuned specific to each task, with the details described below.  

\subsubsection{Testing on an Unknown Environment}

From the 11 sequences in the dataset, 7 were considered for training and 4 for testing. Here, the testing sequences were chosen such that they belonged to different environmental conditions as compared to the training sequences. The network architecture consists of two parallel AlexNet-based cascaded convolutional layers concatenating at the end of the final convolutional layer to generate fully connected layers, which are smoothly stacked to regress the target variables ($\Delta$x, $\Delta$z, $\Delta\Theta$) (Figure {\ref{fig:CNN-2}). 
\begin{figure}[htbp!]
\includegraphics[width=8.0 cm]{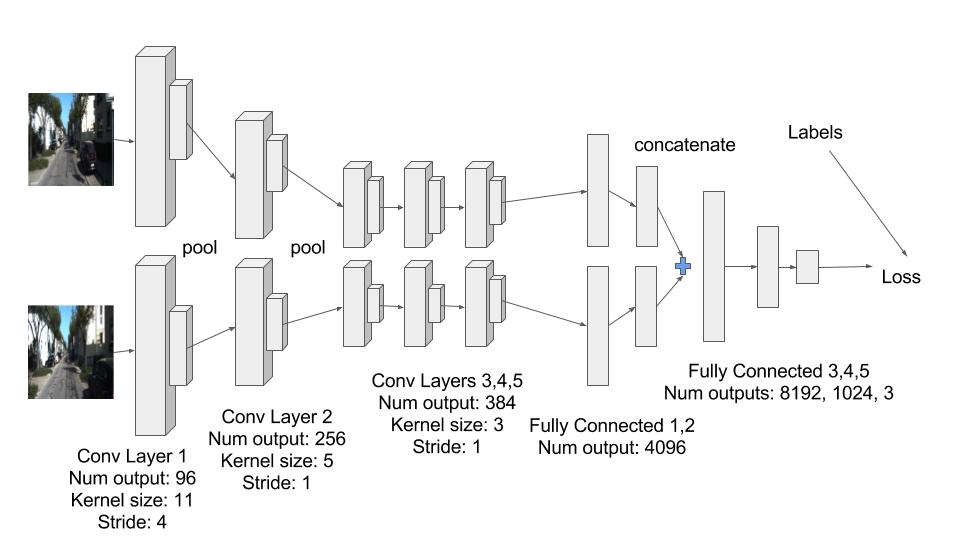}
\caption{Architecture used for Unknown Case}
\label{fig:CNN-2}
\end{figure}
\\
The network takes 3 inputs in the form of I$_{t}$, I$_{t+1}$ and the pose ($\Delta$x, $\Delta$z, $\Delta\Theta$) between them. The two data inputs corresponding to image sequences were fed into the convolutional cascades which convolved in parallel, and then concatenated at the end to generate a flattened (image batch size x 8192) vector. This vector was fed into custom designed fully connected layers that converged to (image batch size x 3) and was fed along with the ground truth label to an Euclidean loss layer to minimize the loss. The same architecture, ignoring the dropout layers, was used in test phase.

\subsubsection{Testing on a Known Environment}

The training sequences and testing sequence were taken from a random permutation of the entire dataset into two different proportions: 80:20 and 50:50 from all the 11 sequences individually. This ensured that both training and test sets contained similar environment sequences. 

The network architecture adopted was exactly the same as the previous experiment. The only difference from the previous experiment was in the preparation of the training set and testing set, with the motivation to observe the network's behavior in a known or unknown environment. This provides an insight into the nature of the Visual Odometry problem. The experiment helps in understanding if the proposed network architecture is robust to new environments or requires a prior knowledge of the scene. 

The model was trained twice independently, once for the 80:20 and once for 50:50 training to testing set ratio scenario. The major motivation for training the model in two different ratios was to analyze the amount of data required by the network to sufficiently learn about the environment to be able to accurately estimate the trajectory.

\subsubsection{Testing on an unknown environment with prior features}

\begin{figure}[htbp!]
\begin{subfigure}{.25\textwidth}
 \centering
 \includegraphics[scale=0.35]{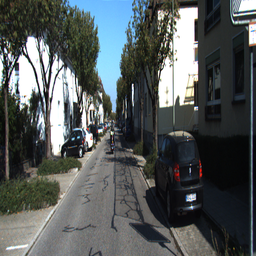}
 \caption{Original Image}
\end{subfigure}\hfill%
\begin{subfigure}{.20\textwidth}
 \centering
 \includegraphics[scale=0.35]{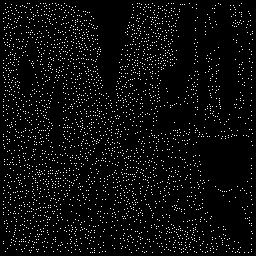}
 \caption{FAST features}
\end{subfigure}%
\caption{Representation of FAST features in the network}
\label{fig:FAST}
\end{figure}

For this task, in addition to the schema used in the first experiment, FAST \cite{rosten2006machine} features were added as a prior input to the network (Figure \ref{fig:FAST}). The features for each image were appended to the RGB data to generate a 4-dimensional feature set for the each input image. The image data thus obtained and the poses ground truth were segregated into 7 training and 4 test sequences. The network architecture, same as the previous experiments, follows the the same procedure as employed in the first experiment. This experiment was performed with an objective to observe the influence of a prior feature, conventionally used for a feature-based approach for solving the visual odometry problem, in improving the accuracy of pose estimation. 

\subsubsection{Testing on an unknown environment using pre-trained network.}

\begin{figure}[htbp!]
\includegraphics[width=8.0 cm]{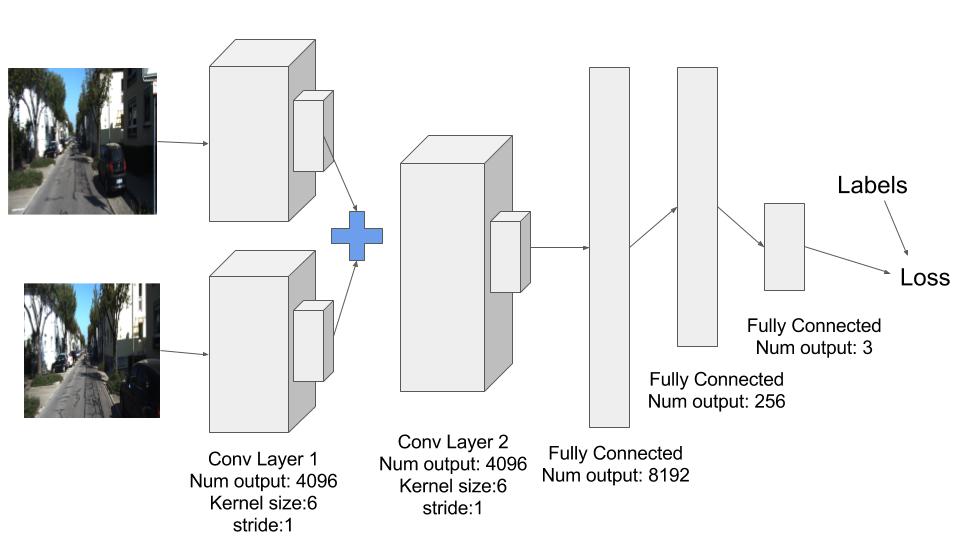}
\caption{AlexNet-based architecture for unknown environment with pre-trained network}
\label{fig:CNN-3}
\end{figure}

This experiment was performed using a network architecture consisting of two AlexNet-based cascaded convolutional layers pre-trained on the ImageNet database. The network was fine-tuned by training on part of dataset sequences while the rest were used as test sequences. Here, the output activations of the final convolutional layer in the original AlexNet architecture were extracted and served as the input instead of a standard RGB image. The learnable part of the architecture comprised of 1 convolution layer and 4 fully connected layers (Figure \ref{fig:CNN-3}). This experiment was designed with the motivation to understand the effect of pre-trained activations trained on object classification labels for the task of estimating the odometry vector.

\section{Experimental Results}
For the experiments described in section 3.3, the results are shown for comparison of the network predictions with the ground truth and to observe the loss in training and testing phase. The network was observed to pass any arbitrary image pair through its layers, compute the layer activations and estimate the odometry vectors at an average of 9ms, displaying real-time capabilities. It was further observed that this did not depend on the nature of the scene.   

\subsection{Test Results : Unknown Environment}
For this evaluation, the testing was performed on an environment completely unknown to the network. In such conditions, the estimated position deviates too much from the ground truth, as shown in Figure \ref{Exp1}.
\begin{figure}[htbp!]
\includegraphics[width=8.0 cm]{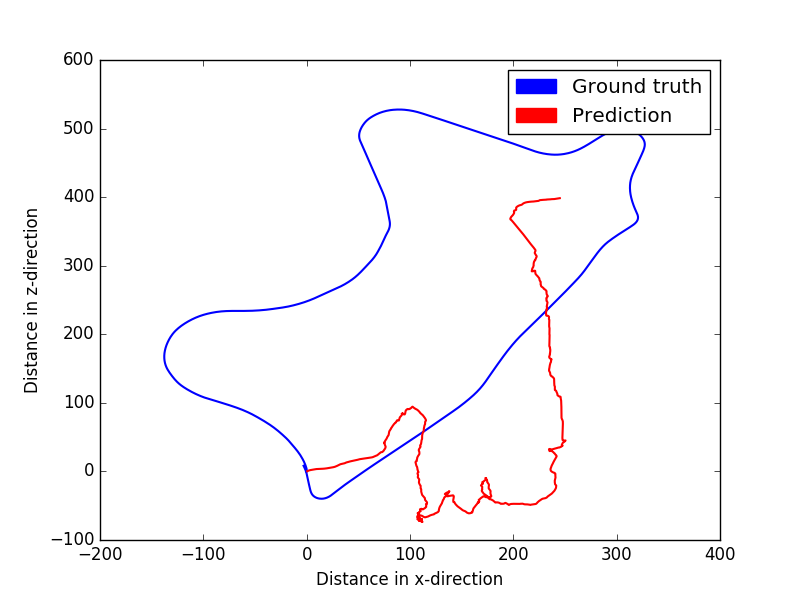}
\caption{Prediction vs Ground Truth: Unknown Environment}
\label{Exp1}
\end{figure}
The training and test loss for this network is shown in Figure \ref{Exp1_loss}. As can be observed from the plot, the training loss declines very fast with the number of iterations. On the other hand, the loss during testing oscillates around a fixed value with small variations. This shows that although the network is able to reduce the the loss on a known environment, the lack of knowledge of a scene does not help in estimating the odometry vector. Therefore, even after a significant number of iterations, the testing loss does not fall.

\begin{figure}[htbp!]
\includegraphics[width=8.0 cm]{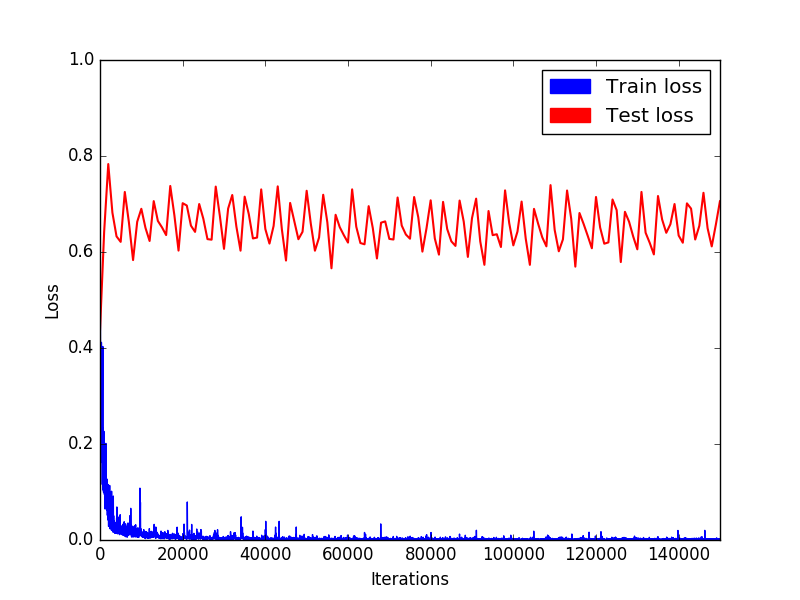}
\caption{Training and Testing loss : Unknown Environment}
\label{Exp1_loss}
\end{figure}

\subsection{Test Results : Known Environment}
This experiment was performed on a known environment, with data segregated into training and testing sequence in ratios of 80-20 and 50-50. Figure \ref{Exp2_path_11} and \ref{Exp2_path_41} show a significant improvement in the prediction of odometry vector in a sequence, part of which is already known to the network. Figure \ref{Exp2_path_11}, \ref{Exp2_dev_11} and \ref{Exp2_loss_11} are the results for data broken into 50-50 ratio.

\begin{figure}[htbp!]
\includegraphics[width=8.0 cm]{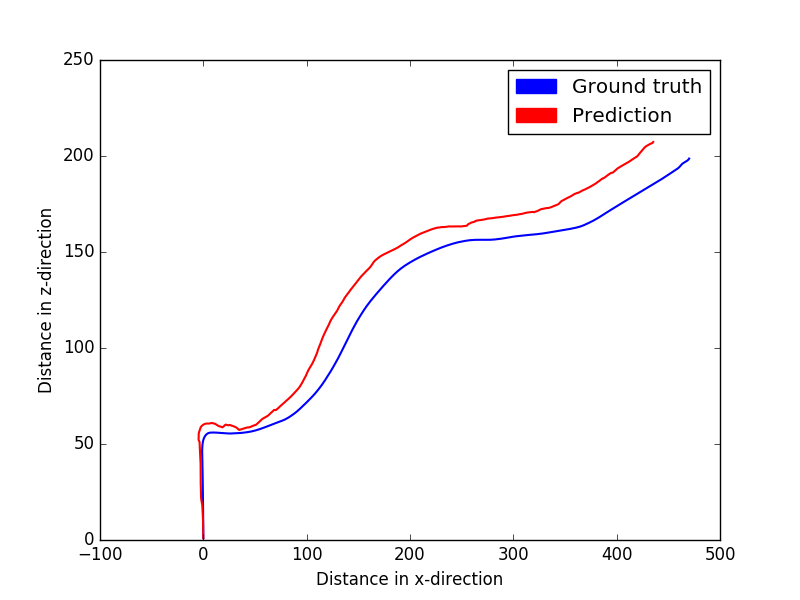}
\caption{Comparison of the predicted output with the ground truth (50-50 proportion of training and test data) : Known Environment}
\label{Exp2_path_11}
\end{figure}

Figure \ref{Exp2_dev_11} gives an insight into the deviation, which is observed to be increasing with time. Therefore, it can be concluded that the error in odometry accumulates over time resulting in the predicted trajectory drifting away from the ground truth.

\begin{figure}[htbp!]
\includegraphics[width=8.0 cm]{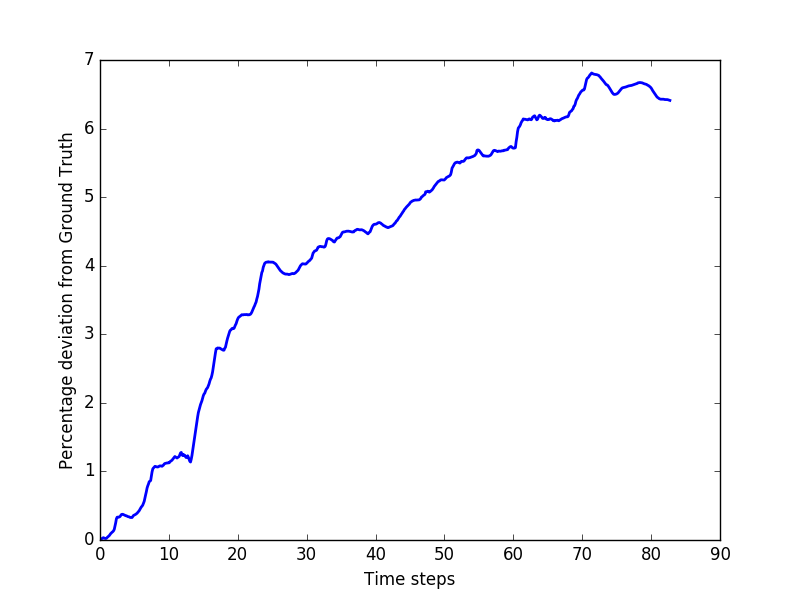}
\caption{Deviation from the ground truth for test in known environment (50-50 proportion of training and test data) : Known Environment}
\label{Exp2_dev_11}
\end{figure}
The loss, similar to deviation, shows great improvement in performance in known environment over unknown environment. The test loss follows the training loss and shows a steep drop with increase in number of iterations.
\begin{figure}[htbp!]
\includegraphics[width=8.0 cm]{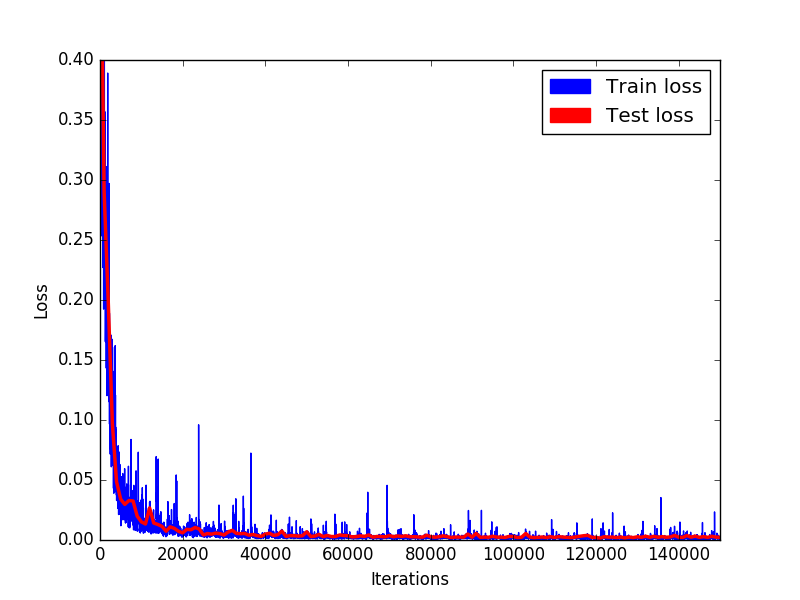}
\caption{Training and testing loss for test in known environment(50-50 proportion of training and test data)}
\label{Exp2_loss_11}
\end{figure}

Figure \ref{Exp2_path_41}, \ref{Exp2_dev_41} and \ref{Exp2_loss_41}  depict the result with same methodology but for a separation of data in 80-20 proportions (training and test data).
\begin{figure}[htbp!]
\includegraphics[width=8.0 cm]{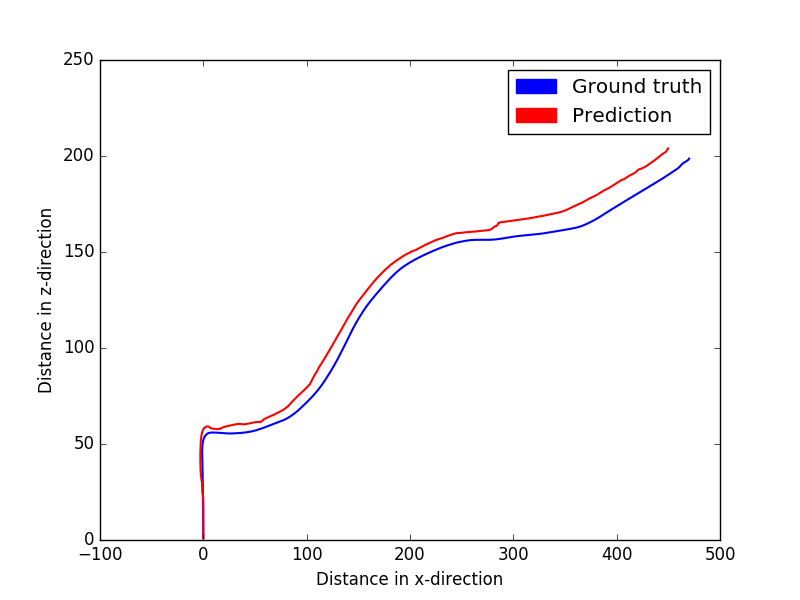}
\caption{Comparison of the predicted output with the ground truth (4:1 ratio of training and test data)}
\label{Exp2_path_41}
\end{figure}

\begin{figure}[htbp!]
\includegraphics[width=8.0 cm]{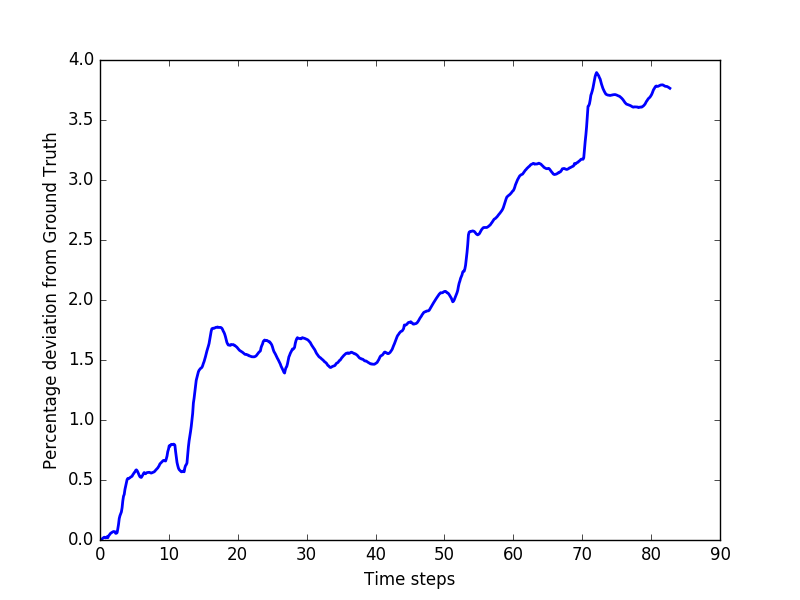}
\caption{Deviation from the ground truth for test in known environment (80-20 proportion of training and test data)}
\label{Exp2_dev_41}
\end{figure}

\begin{figure}[htbp!]
\includegraphics[width=8.0 cm]{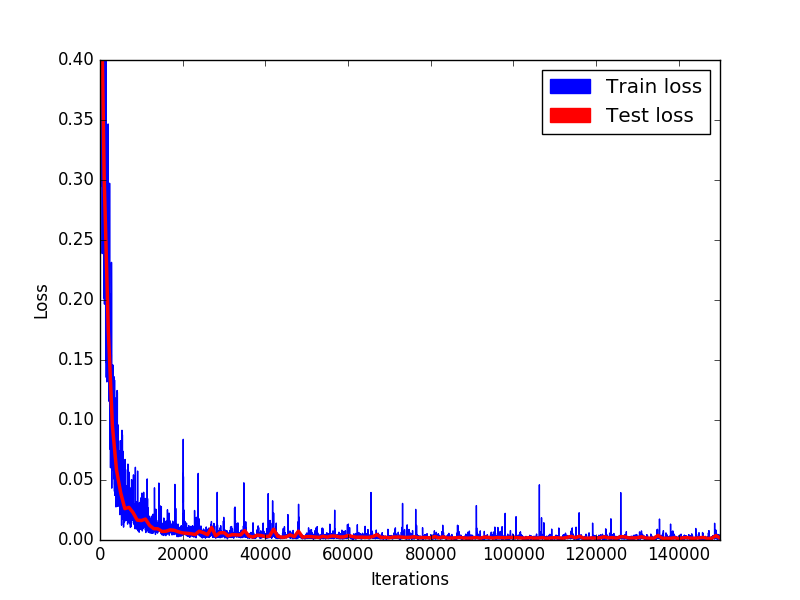}
\caption{Training and testing loss for test in known environment (80-20 proportion of training and test data)} 
\label{Exp2_loss_41}
\end{figure}

\begin{figure}[htbp!]
\includegraphics[width=8.0 cm]{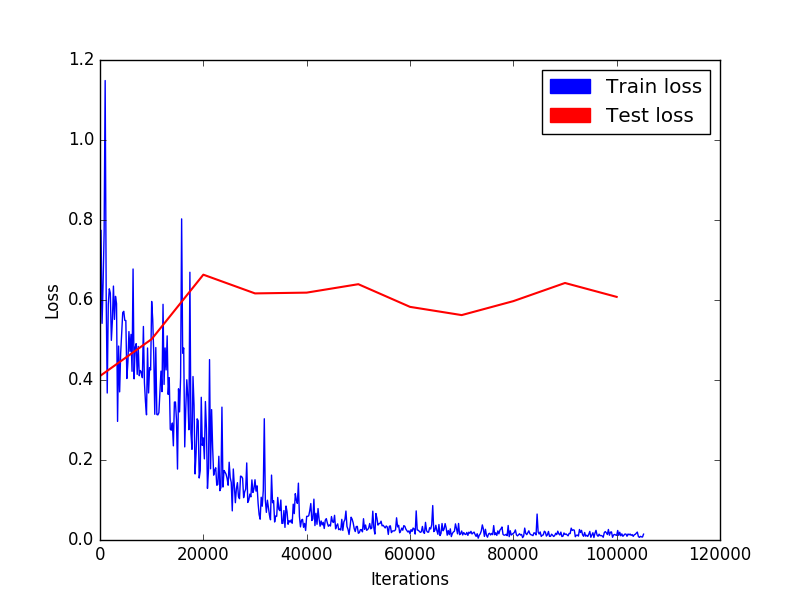}
\caption{Training and testing loss for test in unknown environment with prior features} 
\label{fig4_fast}
\end{figure}

\subsection{Test Results : Using a trackable prior feature in an Unknown Environment}

In this part, we used FAST features as priors along with the RGB images. As observed from Figure \ref{fig4_fast}, this network displays similar behavior in terms of training and test loss as that of a network in an unknown environment. This experiment consisted of fewer test iteration cases.

\section{Discussions}

The results from the experiments performed are highly encouraging. The authors believe that the results not only suggest that the architecture presented can be tried out on robotic platforms, but also provide us a deep understanding of how this network deals with the visual odometry problem. 

From the results of testing on a known environment, it is clear that more the network learns about a particular environment, the better it gets at predicting the visual odometry. This is in alignment with the general perception. Also, this supports the hypothesis that the network treats the problem of visual odometry as specific to a particular scene. This is further supported on comparing these results to that of 1$^{st}$ experiment. In case of predicting visual odometry data on unseen images, the network performs fairly poor. 

Inspired by this finding, the authors delve deeper into understanding the significance of features required for scene understanding. \cite{agrawal2015learning} presents the use of ego-motion vector as a weak supervisory signal for feature learning. They show the effectiveness of the features learnt on simple tasks like scene and object recognition. Motivated by this, the authors used the pre-trained weights of AlexNet \cite{alexnet} trained on object classification for the presented network. However the results obtained are not supportive of the fact, thus showing that the features extracted from the pre-trained network are not generic to the problem of visual odometry.

The authors try out the idea of providing prior information about the scene to improve the prediction accuracy on unknown environments. Therefore, the FAST features of the scene were used along with the features extracted by the convolutional layers of the network.

\subsection{Future Work}
The results of predicting visual odometry in known environment shows the error drifting with time. Therefore, the predicted trajectory also seems to show more deviation from ground truth with time. To tackle this issue, the authors feel that the use of recurrent network would be more appropriate. The presence of recurrent connections would enable the network to correct the error incurred from ground truth continuously. 

It would also be interesting to explore further on the fusion of conventional trackable features as a prior to the higher level features generated by the CNNs. 

Use of generative networks to predict the next scene from an estimated ego-motion vector and update the ego-motion vector using a feedback loop could be used to correct the accumulating error. The mechanism is known to function in the human brain \cite{synofzik2008beyond} and a similar architecture can be used in artificial systems too.

\section{Conclusions}

The proposed network demonstrates promising results, when provided with a prior knowledge of the environment, while displaying the expected opposite response in case of an unknown environment. The network, when provided with a prior of FAST features, and trained on an unknown environment, shows a similar behavior as that of the network subjected to an unknown environment without any prior. It may be concluded that the proposed CNN designed for the purpose of Visual Odometry is able to learn features similar to FAST, and a manual addition of these features only contributes to redundancy. When deployed on known environments, the network architecture is able to learn the actual scale in real time, which is not possible for monocular visual odometry using geometric methods. 

{\small
\bibliographystyle{ieee}
\bibliography{egbib}
}

\end{document}